\documentclass[final]{cvpr}

\usepackage{times}
\usepackage{epsfig}
\usepackage{graphicx}
\usepackage{amsmath}
\usepackage{amssymb}
\usepackage{booktabs}
\usepackage[]{sidecap}
\usepackage{makecell}
\usepackage{graphics}
\usepackage[font=small]{caption}
\usepackage{xcolor}
\definecolor{purple}{HTML}{8000C8}
\definecolor{pink}{HTML}{DE37D1}
\definecolor{bluelight}{HTML}{73B7F9}

\newcommand\blfootnote[1]{%
\begingroup
\renewcommand\thefootnote{}\footnote{#1}%
\addtocounter{footnote}{-1}%
\endgroup
}

\usepackage[pagebackref=true,breaklinks=true,colorlinks,bookmarks=false]{hyperref}

\def\cvprPaperID{3921} 
\def\confYear{CVPR 2021}

\begin{document}

\title{\vspace{-0.5cm}Learning the Predictability of the Future}

\author{D\'idac Sur\'is*, Ruoshi Liu*, Carl Vondrick\\Columbia University\\\href{http://hyperfuture.cs.columbia.edu}{\textbf{\textcolor{purple}{\texttt{hyperfuture.cs.columbia.edu}}}}
}

\maketitle


\begin{abstract}
    We introduce a framework for learning from unlabeled video what is predictable in the future. Instead of committing up front to features to predict, our approach learns from data which features are predictable. 
    Based on the observation that hyperbolic geometry naturally and compactly encodes hierarchical structure, we propose a predictive model in hyperbolic space. When the model is most confident, it will predict at a concrete level of the hierarchy, but when the model is not confident, it learns to automatically select a higher level of abstraction. Experiments on two established datasets show the key role of hierarchical representations for action prediction. Although our representation is trained with unlabeled video, visualizations show that action hierarchies emerge in the representation.
\end{abstract}

\vspace{-1em}

\section{Introduction}

\blfootnote{*Equal contribution}

\begin{figure}
    \centering
    \includegraphics[trim=0 0 0 0,width=\linewidth]{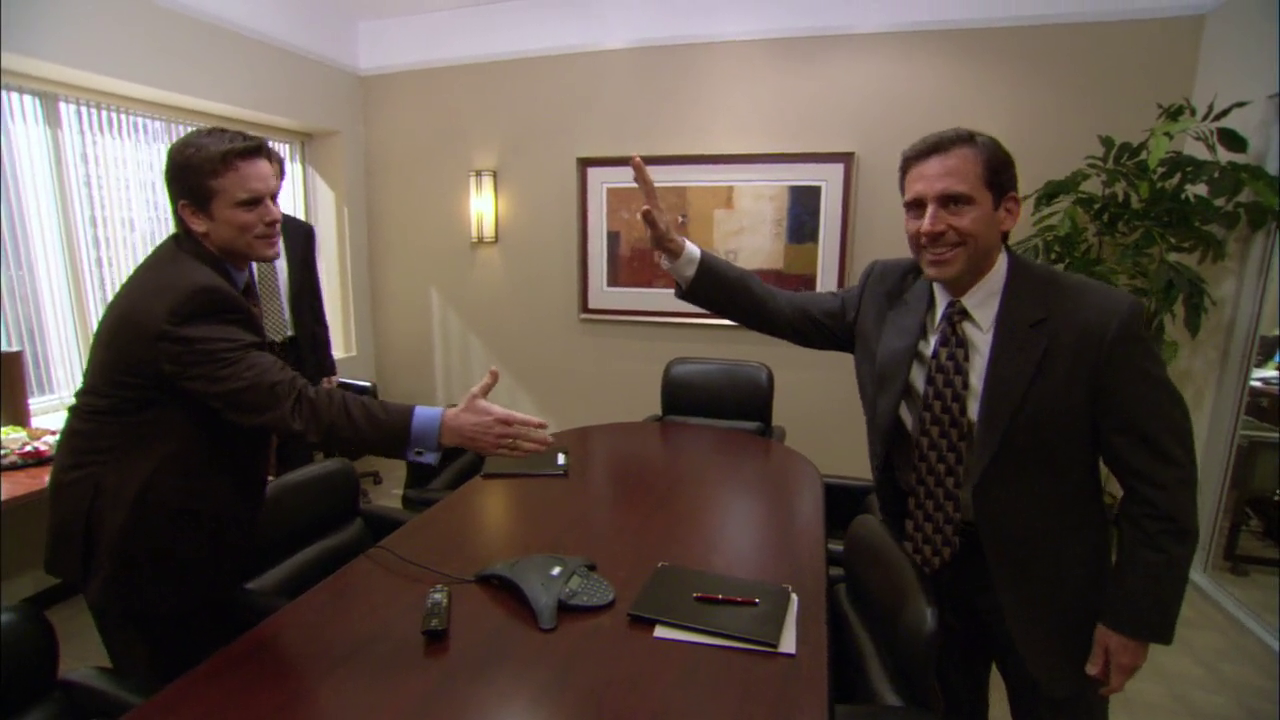} 
    \caption[]{The future is often uncertain. Are they going to shake hands or high five?\footnotemark\ Instead of answering this question, we should ``hedge the bet'' and predict the hyperonym that they will \emph{at least} greet each other.
    In this paper, we introduce a hierarchical predictive model for learning what is predictable from unlabeled video.}
    \label{fig:teaser}
    \vspace{-1em}
\end{figure}

The future often has narrow predictability.
No matter how much you study Fig.~\ref{fig:teaser}, you will not be able to anticipate the exact next action with confidence. Go ahead and study it. Will they shake hands or high five?\footnotemark

For the past decade, predicting the future has been a core computer vision problem  \cite{yuen2010data,kitani2012activity,koppula2015anticipating,walker2014patch,mathieu2015deep,ranzato2014video,fragkiadaki2015learning,vondrick2016anticipating,vondrick2016generating} with a number of applications in robotics, security, and health. Since large amounts of video are available for learning, the temporal structure in unlabeled video provides excellent incidental supervision for learning rich representations \cite{miech20endtoend,fernando2017self,lee2017unsupervised,misra2016shuffle,xu2019self,wang2019self,Sun_2019,han2019video,vondrick2016anticipating,vondrick2018tracking,dwibedi2019temporal,wang2015unsupervised,wang2019learning}. While visual prediction is challenging because it is an underconstrained problem, a series of results in neuroscience have revealed a biological basis in the brain for how humans anticipate outcomes in the future  \cite{kourtzi2000activation,baldassano2017discovering,stachenfeld2017hippocampus}.


However, the central issue in computer vision has been selecting \emph{what} to predict in the future. The field has investigated a spectrum of options, ranging from generating pixels and motions \cite{walker2017pose,kumar2019videoflow,kwon2019predicting} to forecasting activities \cite{abu2018will,vondrick2016anticipating,walker2016uncertain} in the future. However, most representations are not \emph{adaptive} to the fundamental uncertainties in video. While we cannot forecast the concrete actions in Fig.~\ref{fig:teaser} nor generate its motions, all hope is not lost. Instead of forecasting whether the people will high five or shake hands, there is something else predictable here. We can ``hedge the bet'' and predict the abstraction that they will at least greet each other.

This paper introduces a framework for learning from unlabeled video what is predictable. Instead of committing up front to a level of abstraction to predict, our approach learns from data which features are predictable. Motivated by how people organize action hierarchically \cite{barker1955midwest}, we propose a hierarchical predictive representation. Our approach jointly learns a hierarchy of actions while also learning to anticipate at the right level of abstraction.



\setcounter{footnote}{0}

\footnotetext{The answer is in Season 2, Episode 16 of the \emph{The Office}.}

Our method is based on the observation that hyperbolic geometry naturally and compactly encodes hierarchical structure. Unlike Euclidean geometry, hyperbolic space can be viewed as the continuous analog of a tree \cite{Nickel2017} because tree-like graphs can be embedded in finite-dimension with minimal distortion \cite{Gromov1987HyperbolicG}. We leverage this property and learn predictive models in hyperbolic space. When the model is confident, it will predict at the concrete level of the hierarchy, but when the model is not confident, it learns to automatically select a higher-level of abstraction.

Experiments and visualizations show the pivotal role of hierarchical representations for prediction. On two established video datasets, our results show predictive hyperbolic representations are able to both recognize actions from partial observations as well as forecast them in the future better than baselines. 
Although our representation is trained with unlabeled video, visualizations also show action hierarchies automatically emerge in the hyperbolic representation. The model explicitly represents uncertainty by trading off the specificity versus the generality of the prediction. 

The primary contribution of this paper is a hierarchical representation for visual prediction. The remainder of this paper describes our approach and experiments in detail. Code and models are publicly available on \href{https://github.com/cvlab-columbia/hyperfuture}{github.com/cvlab-columbia/hyperfuture}.

\section{Related Work}

\textbf{Video representation learning} aims to learn strong features for a number of visual video tasks. By taking advantage of the temporal structure in video, a line of work uses the future as incidental supervision for learning video dynamics \cite{xingjian2015convolutional, srivastava2015unsupervised, mathieu2015deep}.
Since generating pixels is challenging, \cite{misra2016shuffle, fernando2017self, wei2018learning} instead use the natural temporal order of video frames to learn self-supervised video representations. Similar to self-supervised image representations \cite{doersch2015unsupervised}, temporal context also provides strong incidental supervision \cite{isola2015learning, wang2015unsupervised}.

A series of studies from Oxford has investigated how to \emph{learn} a representation in the future \cite{han2019video, han2020memory} using a contrastive objective. We urge readers to read these papers in detail as they are the most related to our work. While these models learn predictable features, the underlying representation is not adaptive to the varying levels of uncertainty in natural videos. They also focus on action recognition, and not action prediction. By representing action hierarchies in hyperbolic space, our model has robust inductive structure for hedging uncertainty. 

Unlike action recognition, \textbf{future action prediction} \cite{kitani2012activity} and \textbf{early action prediction} \cite{ryoo2011human,hoai2014max} are tasks with an intrinsic uncertainty caused by the unpredictability of the future. Future action prediction infers future actions conditioned on the current observations. Approaches for future prediction range from classification \cite{ng2019forecasting,abu2018will}, through prediction of future features \cite{vondrick2016anticipating}, to generation of the future at the skeletal \cite{li2018convolutional} or even pixel \cite{gammulle2019predicting,jayaraman2018time} levels. Early action prediction aims to recognize actions before they are completely executed. While standard action recognition methods can be used for this task \cite{raptis2013poselet,vahdat2011discriminative}, most approaches mimic a sequential data arrival \cite{kong2017deep,kong2014discriminative,wang2019progressive}. We evaluate our self-supervised learned representations on these two tasks.

\textbf{Hyperbolic embeddings} have emerged as excellent hierarchical language representations in natural language processing \cite{Nickel2017, Ifrea2019}. These works are pioneering.  Riemmanian optimization algorithms \cite{Bonnabel2013, becigneul2018riemannian} are used to optimize the models using hyperbolic geometry. Their success is largely attributed to the advantage of hyperbolic space to represent hierarchical structure. Following the Poincare embedding \cite{Nickel2017}, \cite{Ganea2018a} use a hyperbolic entailment cone to represent the hierarchical relation in an acyclic graph. \cite{Ganea2018} further applies hyperbolic geometry to feedforward neural networks and recurrent neural networks.

Since visual data is naturally hierarchical, hyperbolic space provides a strong inductive bias for images and videos as well. \cite{Khrulkov2019, Dhall2020, Liu2020} perform several image tasks, demonstrating the advantage of hyperbolic embeddings over Euclidean ones. \cite{Long2020} proposes video and action embeddings in the hyperbolic space and trains a cross-modal model to perform hierarchical action search. We instead use hyperbolic embeddings for prediction and we use the hierarchy to model uncertainty in the future. We also learn the hierarchy from self-supervision, and our experiments show an action hierarchy emerges automatically. 


Since dynamics are often stochastic, \textbf{uncertainty representation} underpins predictive visual models.
There is extensive work on probabilistic models for visual prediction, and we only have space to briefly review. 
For example, \cite{Henaff2019} measures the covariance between outputs generated under different dropout masks, which reflects how close the predicted state is to the training data manifold. A high covariance indicates that the model is confident about its prediction. \cite{Balaji2017} use ensembling to estimate uncertainty, grounded in the observation that a mixture of neural networks will produce dissimilar predictions if the data point is rare. Another line of work focuses on generating multiple possible future prediction, such as variational auto-encoders (VAE) \cite{Kingma2014,walker2016uncertain}, variational recurrent neural networks (VRNN) \cite{Chung2015,castrejon2019improved} and adversarial variations \cite{Lee2018a}. These models allow sampling from the latent space to capture the multiple outcomes in the output space.

Probabilistic approaches are compatible with our framework. The main novelty of our method is that we represent the future uncertainty hierarchically in a hyperbolic space. The hierarchy naturally emerges during the process of learning to predict the future. 

\section{Method}

We present our approach for learning a hierarchical representation of the future. We first discuss our model, then introduce our hyperbolic representation.

\subsection{Predictive Model}

Our goal is to learn a video representation that is predictive of the future. Let $x_t \in \mathbb{R}^{T \times W \times H \times 3}$ be a video clip centered at time $t$.  Instead of predicting the pixels in the future, we will predict a representation of the future.
We denote the representation of a clip as $z_t = f(x_t)$.


The prediction task aims to forecast the unobserved representation $z_{t+\delta}$ that is $\delta$ clips into the future. Given a temporal window of previous clips, our model estimates its prediction of $z_{t+\delta}$ as:
\begin{align}
    \hat{z}_{t+\delta} = \phi(c_t, \delta) \quad \textrm{for} \quad c_t = g(z_1, z_2, \ldots, z_t)
\end{align}
where $c_t = g(\cdot)$ contextually encodes features of the video from the beginning up to and including frame $t$.

We will model $f$, $g$, and $\phi$ each with a neural network. In order to learn the parameters of these models, we need to define a distance metric between the unobserved future $z$ and the prediction $\hat{z}$.
However, the future is often uncertain. Instead of predicting the exact $z_{t+\delta}$ in the future, our goal is to predict an abstraction that encompasses the variation of the possible futures, and no more. 




\subsection{Hierarchical Representation}

The key contribution of this paper is to predict a hierarchical representation of the future.
When the future is certain, our model should predict $z_{t+\delta}$ as specifically as possible. However, when the future is uncertain, our model should ``hedge the bet'' and forecast a hierarchical parent of $z_{t+\delta}$. For example, in Fig.~\ref{fig:teaser} the parent of a hand shake and a high five is a greeting.
In order to parameterize this hierarchy, we will learn predictive models in \emph{hyperbolic} space.

Informally, hyperbolic space can be viewed as the continuous analog of a tree \cite{Nickel2017}.
Unlike Euclidean space, hyperbolic space has the unique property that circle areas and lengths grow exponentially with their radius. This density allows hierarchies and trees to be compactly embedded \cite{Gromov1987HyperbolicG}.
Since this space is naturally suited for hierarchies, hyperbolic predictive models are able to smoothly interpolate between forecasting abstract representations (in the case of low predictability) to concrete representations (in the case of high predictability). 

\begin{figure}
\includegraphics[width=\linewidth]{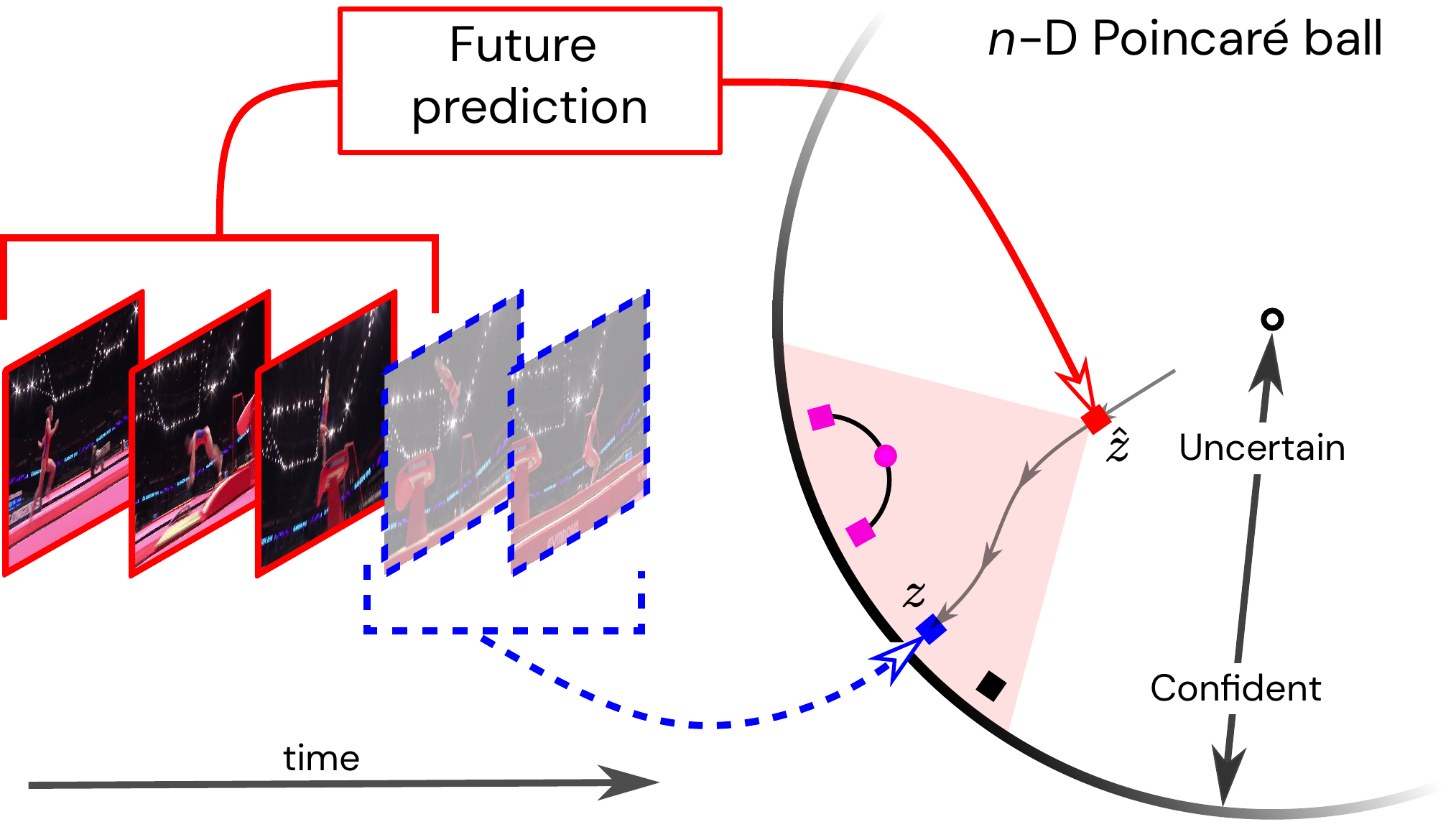}
\caption{The future is non-deterministic. Given a specific past (first three frames in the figure), different representations (represented by squares in the Poincar\'e ball) can encode different futures, all of them possible. In case the model is uncertain, it will predict an abstraction of all these possible futures, represented by $\hat{z}$ (\textcolor{red}{red} square). The more confident it is, the more specific the prediction can get. Assuming the actual future is represented by $z$ (\textcolor{blue}{blue} square), the gray arrows represent the trajectory the prediction will follow as more information is available. The \textcolor{pink}{pink} circle exemplifies the increase in generality when computing the mean of two specific representations (\textcolor{pink}{pink} squares).}
\label{fig:poincare_ball}
\vspace{-1em}
\end{figure}

The hyperbolic $n$-space, which we denote as $\mathbb{H}^n$, is a Riemannian geometry that has constant negative curvature.\footnote{We assume the curvature to be $-1$. Hyperbolic is one of three isotropic model spaces. The other two spaces are the Euclidean space $\mathbb{R}^n$ (zero curvature), and the spherical space $\mathbb{S}^n$ (constant positive curvature).} 
While there are several models for hyperbolic space, we will use the the Poincar\'e model, which is also the most commonly used in gradient-based learning. The Poincar\'e ball model is formally defined by the manifold $\mathbb{D}^n = \{X\in\mathbb{R}^n : ||x|| < 1\}$ and the Riemmanian metric $g^\mathbb{D}$:
 $
  g_{x}^{\mathbb{D}}=\lambda_{x}^{2} g^{E} $ where $ \lambda_{x}:=\frac{2}{1-\|x\|^{2}} 
  $
such that $g^E=\textbf{I}_n$ is the Euclidean metric tensor. For more details, see \cite{lee2006riemannian,lee2013smooth}.

We use the Poincar\'e ball model to define the distance metric between a prediction $\hat{z}$ and the observation $z$:
\begin{equation}
    d_{\mathbb{D}}(\hat{z}, z)=\cosh ^{-1}\left(1+2 \frac{\|z-\hat{z}\|^{2}}{\left(1-\|z\|^{2}\right)\left(1-\|\hat{z}\|^{2}\right)}\right)
    \label{eq:distance}
\end{equation}
for points on the manifold $\mathbb{D}^n$. Recall that the mean minimizes the sum of squared residuals.  The key property is that, 
in hyperbolic space, the mean between two leaf embeddings is not another leaf embedding, but an embedding that is a parent in the hierarchy. If the model cannot select between two leaf embeddings given the provided information, the expected squared distance will be minimized by instead producing the more abstract one as the prediction.

Fig.~\ref{fig:poincare_ball} visualizes this property on the Poincar\'e ball. Points near the center of the ball (having a smaller radius) represent abstract embeddings, while points near the edge (having a large radius) represent specific ones. In this example, the mean of two points close to the edge of the ball---illustrated by the two \textcolor{pink}{pink} squares in Fig.~\ref{fig:poincare_ball}---is a node further from the edge, represented with a pink circle. The line connecting the two squares is the minimum distance path between them, or \textit{geodesic}. The midpoint is the mean.

Unlike trees, hyperbolic space is continuous, and there is not a fixed number of hierarchy levels. Instead, there is a continuum from very specific (closer to the border of the Poincar\'e ball) to very abstract (closer to the center).



\subsection{Learning}

To learn the parameters of the model, we want to minimize the distance between the predictions $\hat{z}_t$ and the observations $z_t$. We use the contrastive learning objective function \cite{han2019video} with hyperbolic distance as the similarity measure: 
\begin{equation}
    \mathcal{L}=-\sum_{i}\left[\log \frac{\exp \left(-d^2_{\mathbb{D}}({{}\hat{z}}_i, {z}_{i})\right)}{\sum_{j} \exp \left(-d^2_{\mathbb{D}}({{}\hat{z}}_i, {z}_{j})\right)}\right]
    \label{eq:loss}
\end{equation}

where $z$ is the feature representing a spatio-temporal location in a video, and $\hat{z}$ is the prediction of that feature. The contrastive objective pulls positive pairs $z$ and $\hat{z}$ together while also pushing $\hat{z}$ away from a large set of negatives, which avoids the otherwise trivial solution. 
There are a variety of strategies for selecting negatives \cite{chen2020simple,tian2019contrastive,han2019video}. We create negatives from other videos in the mini batch, as well as using features that correspond to the same video but in different spatial or temporal locations.



The solution to the hyperbolic contrastive objective minimizes the distance between the positive pair $d^2_{\mathbb{D}}({{}\hat{z}}_i, {z}_{i})$. When there is no uncertainty, the loss is minimized if $\hat{z}_t = z_t$. However, in the face of uncertainty between two possible outcomes $a$ and $b$, the loss is minimized by predicting the midpoint on the geodesic between $a$ and $b$. Since hyperbolic space has constant negative curvature, this solution corresponds to the latent parent embedding of $a$ and $b$.


Our approach treats the hierarchy as latent. Since hyperbolic space is continuous and differentiable, we are able to optimize Eq. \ref{eq:loss} with stochastic gradient descent, which jointly learns the predictive models with the hierarchy. The model will learn a hierarchy that is organized around the predictability of the future. 



\subsection{Classification}

After the representation $z$ is trained, we are able to fit any classifier on top of it.
We use a linear classifier, and keep the rest of the representation fixed (no fine-tuning). However, since the representation is hyperbolic, we cannot use a standard Euclidean linear classifier. Instead, we use a hyperbolic multiclass logistic regression \cite{Ganea2018} that assumes the input representations are in the Poincar\'e ball, and fits a hyper-plane in the same space. For the Euclidean baseline we train a standard Euclidean multiclass logistic regression. In both models, we treat each node as an independent category---not requiring a ground truth hierarchy---when training the classifier, and then compute accuracy values independently for each hierarchy level. 

\subsection{Network Architecture}

\begin{figure}
  \includegraphics[width=\linewidth]{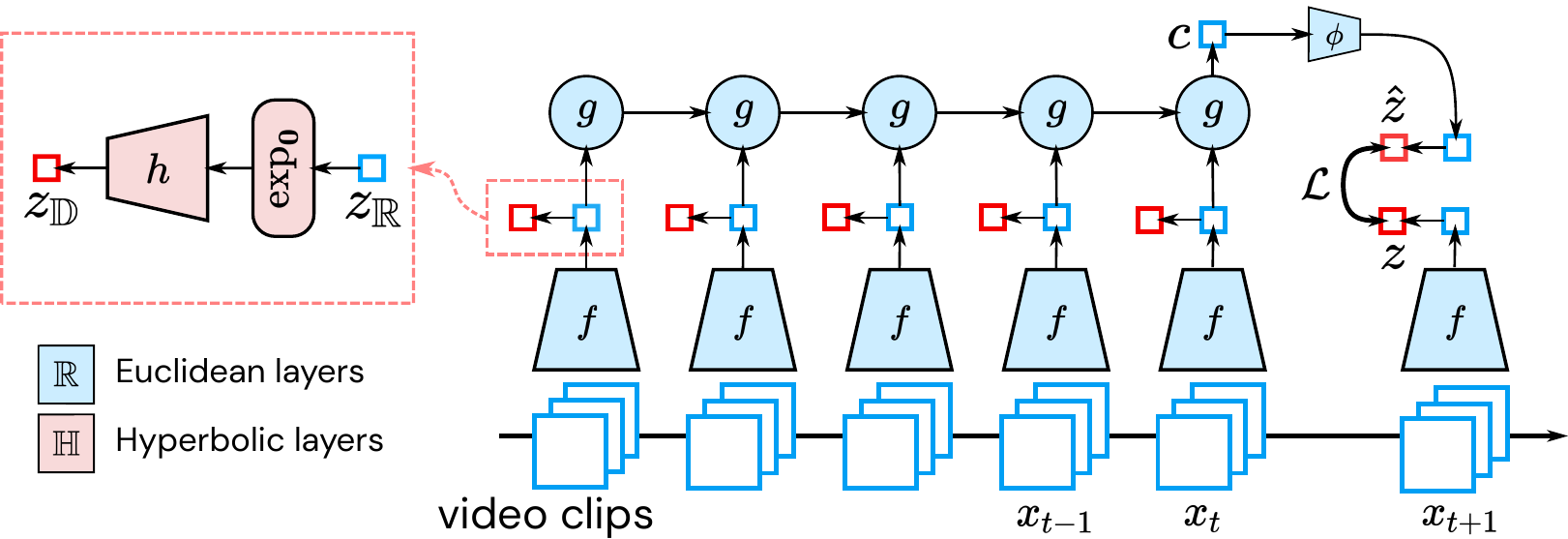}
  \caption{Overview of architecture. Blue and red respectively indicate Euclidean and hyperbolic modules.}
  \label{fig:architecture}
  \vspace{-1em}
\end{figure}

While we estimate our predictions and loss in hyperbolic space, the entire model does not need to be hyperbolic. This flexibility enables us to take advantage of the extensive legacy of existing neural network architectures and optimization algorithms that have been highly tuned for Euclidean space.  We therefore parameterize $f$, $g$, and $\phi$ with neural networks, which will be in Euclidean space. Our approach only instantiates $z$ and $\hat{z}$ in hyperbolic space.
Fig.~\ref{fig:architecture} illustrates this architecture.

\begin{figure*}[t]
\center
  \includegraphics[width=0.9\textwidth]{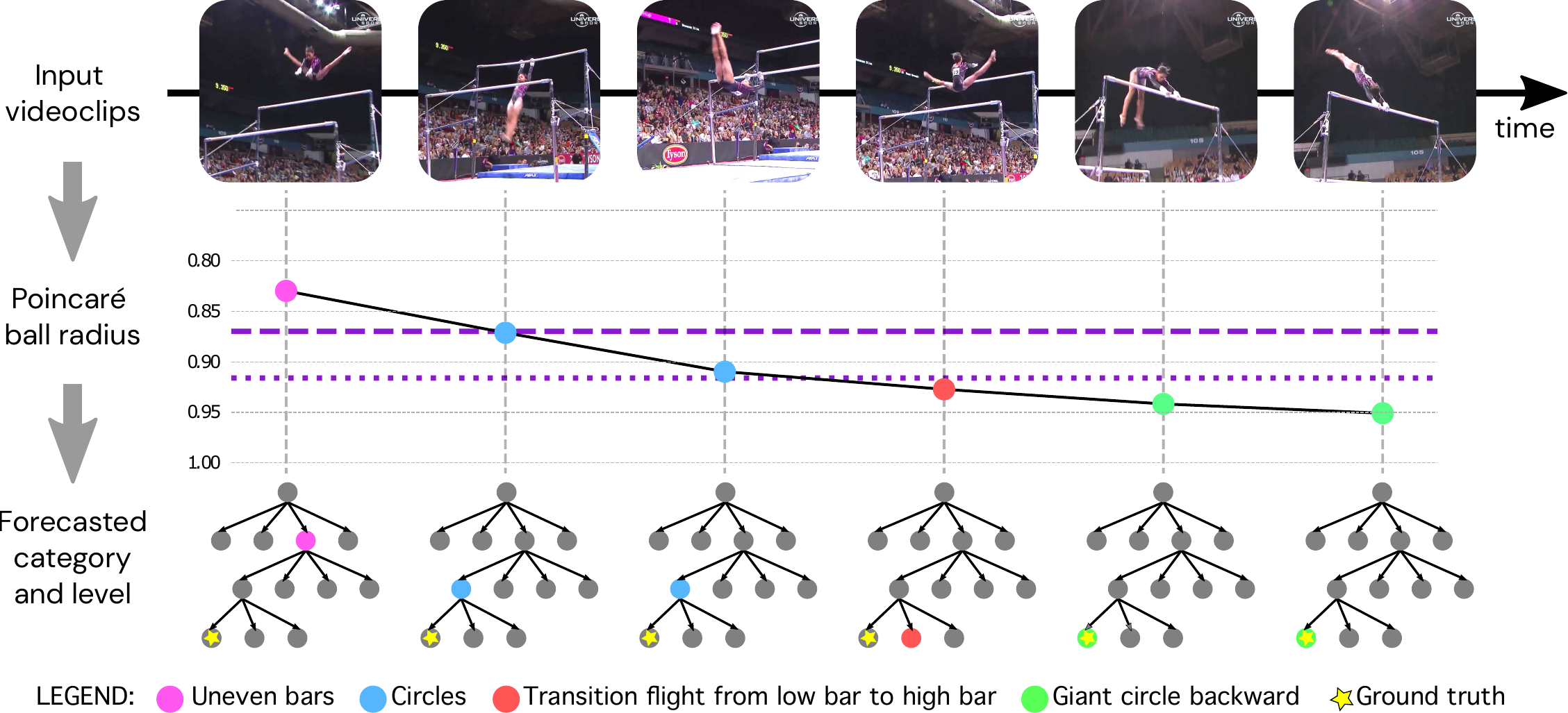}
  \caption{We show an example of future action prediction, where the model has to determine both a hierarchy level and a class within that level. At each time step, the model has to predict the class of the last action in the video. At the top of the figure we show the input video, where each image represents a video clip. From the input video, the model computes a representation and a level in the hierarchy based on the Euclidean norm of this representation. The thresholds shown in \textcolor{purple}{purple} are the 33\% percentile (dashed line) and the 66\% percentile (dotted line), ranked by the radius of predicted hyperbolic embeddings of all videos in FineGym test set. Once a level in the tree has been determined, the model predicts the class within that level. We can see how the closer we get to the actual action to be predicted, the more confident the model is. Also, we can see how being overly confident (by setting a low threshold and choosing a level that is too specific) may lead the model to predict the wrong class. In this specific case, the fourth prediction (\textcolor{red}{red} dot) would have been more accurate had the model predicted a class in the parent level (i.e. had it predicted the \textcolor{bluelight}{blue} dot). Note that the y-axis is inverted in the graph. Also, the nodes represented in the tree are not all the classes in the FineGym dataset, and each node has more children not shown.}
  \label{fig:tree_finegym}
  \vspace{1em}
  \includegraphics[width=\textwidth]{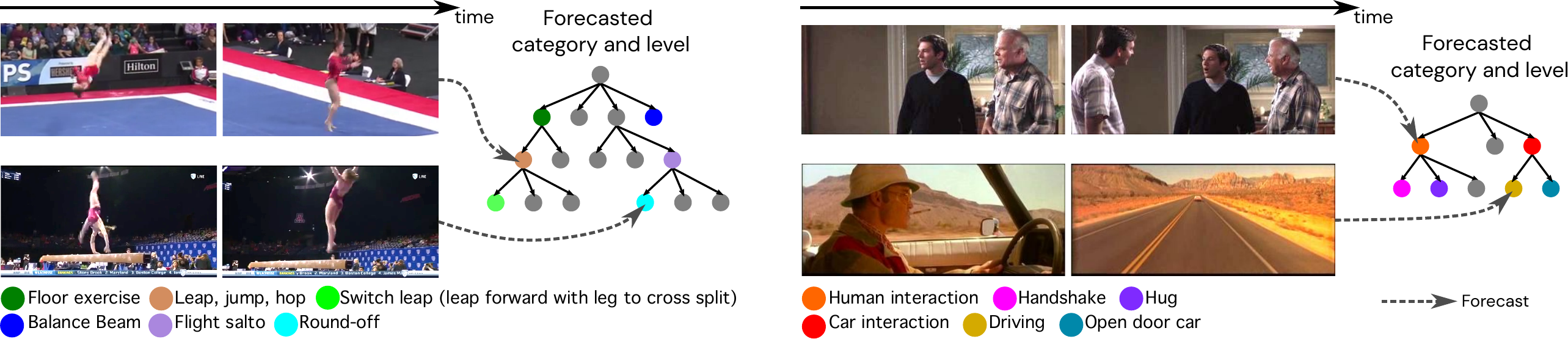}
  \caption{We show four more examples where the model correctly predicts the class at different hierarchical levels. The FineGym examples (left) are evaluated on future action prediction, and the Hollywood2 ones (right) on early action prediction, where we show the model half of the clips in the video. The shown trees are just partial representations of the complete hierarchies. Click \href{https://hyperfuture.cs.columbia.edu/index.html\#results}{here} for video visualizations.}
  \label{fig:more_examples}
  \vspace{-1em}
\end{figure*}

In order to use this hybrid architecture, we need a projection between the two spaces. 
The transition from the Euclidean space is based on the process mapping to Riemannian manifolds from their corresponding tangent spaces. A \textit{Riemnnian manifold} is a pair ($\mathcal{M}$, $g$), where $\mathcal{M}$ is a smooth manifold and $g$ is a Riemmanian metric.
Broadly, smooth manifolds are spaces that locally approximate Euclidean space $\mathbb{R}^n$, and on which one can differentiate \cite{lee2013smooth}, and this is precisely the connection between the two spaces. For $x \in \mathcal{M}$, one can define the tangent space $T_x\mathcal{M}$ of $\mathcal{M}$ at $x$ as the first order linear approximation of $\mathcal{M}$ around $x$. 

The \textit{exponential map} $\text{exp}_x: T_x\mathcal{M}\rightarrow \mathcal{M}$ at $x$ is a map from the tangent spaces into the manifold. This projection is important for several operations, such as performing gradient updates \cite{Bonnabel2013}. The inverse of the exponential map is called \textit{logarithmic map}, denoted $\text{log}_x$. 


We use an exponential map centered at $\textbf{0}$ to project from the Euclidean space to the hyperbolic space \cite{Liu2020}. Once the representations are in the hyperbolic space, the mathematical operations and the optimization follow the rules derived by the metric in that space. 
Essentially, a Riemannian metric defines an inner product $g_x$ that allows us to define a global distance function as the infimum of the lengths of all curves between two points $x$ and $y$ \cite{lee2006riemannian}:
\begin{equation}
    d(x, y)=\inf _{\gamma} L_g(\gamma),
\end{equation}
where $\gamma: [0, 1] \rightarrow \mathcal{M}$ is a curve, and $L_g(\gamma)$ is the length of the curve, defined as:
\begin{equation}
    L_g(\gamma)=\int_{0}^{1}|\dot{\gamma}(t)| d t = \int_{0}^{1} \sqrt{g_{\gamma(t)}(\dot{\gamma}(t), \dot{\gamma}(t))} d t.
\end{equation}

In the specific case of the Poincar\'e ball model, this distance is the same as Eq.~\ref{eq:distance}. Based on these concepts, several papers define extensions of the standard (Euclidean) neural network layers to the hyperbolic geometry \cite{Ganea2018,shimizu2020hyperbolic,Chami2019,Liu2019b,Gulcehre2019}. We use the hyperbolic feed-forward layer defined in \cite{Ganea2018} to obtain the representation $z_\mathbb{D}$ that we use in Eq.~\ref{eq:loss}, from the Euclidean representation $z_\mathbb{R}$. Specifically, we apply this layer after the exponential map, as shown in Fig~\ref{fig:architecture}. If the space is not specified, $z$ is assumed to be $z_\mathbb{D}$.

We implement $f$ with a 3D-ResNet18, $g$ as a one-layer Convolutional Gated Recurrent Unit (ConvGRU) with kernel size $(1,1)$, and $\phi$ using a two-layer perceptron. The dimensionality of the ResNet output is $256$. When training with smaller dimensionality, we add an extra linear layer to project the representations. For more implementation details, we refer the reader to the Appendix \ref{sec:appendix_impl_details}.






\section{Experiments}

The basic objective of our experiments is to analyze how hyperbolic representations encode varying levels of uncertainty in the future. We quantitatively evaluate on two different tasks and two different datasets. We also show several visualizations and diagnostic analysis. 


\subsection{Datasets and Common Setup}


We use a common evaluation setup throughout our experiments. We first learn a self-supervised representation from a large collection of unlabeled videos by optimizing Eq.~\ref{eq:loss}. After learning the representation, we transfer these representations to the target domain using a smaller, labeled dataset. On the target domain, we fine-tune on the same objective before fitting a supervised linear classifier on $\hat{z}$ using a small number of labeled examples.

We evaluate on two different video datasets, which we selected for their realistic temporal structure: 


\textbf{Sports Videos}: In this setting, we learn the self-supervised representation on Kinetics-600 \cite{kay2017kinetics} and fine-tune and evaluate on FineGym \cite{shao2020finegym}. Kinetics has $600$ human action classes and $500,000$ of videos which contain rich and diverse human actions. We discard its labels. FineGym is a dataset of gymnastic videos where clips are annotated with three-level hierarchical action labels, ranging from specific exercise names in the lowest level to generic gymnastic routines (e.g. \textit{balance-beam}) in the highest one. The highest level of the hierarchy is consistent for all the clips in a video, so we use it as a label for the whole video.

\textbf{Movies}: In our second setting, we learn the self-supervised representation on MovieNet \cite{huang2020movienet}, then fine-tune and evaluate the Hollywood2 dataset \cite{marszalek09}. MovieNet contains $1,100$ movies and $758,000$ key frames. In order to obtain a hierarchy of actions from Hollywood2 at the video level, we grouped action classes into more general ones to form a 2-level hierarchy. The hierarchy is shown in the Appendix \ref{sec:appendix_hollywood}. Since Hollywood2 does not have fine-grained clip-level action labels like FineGym, we do not evaluate future action prediction on this dataset.

\begin{figure}
\center
\includegraphics[width=\linewidth]{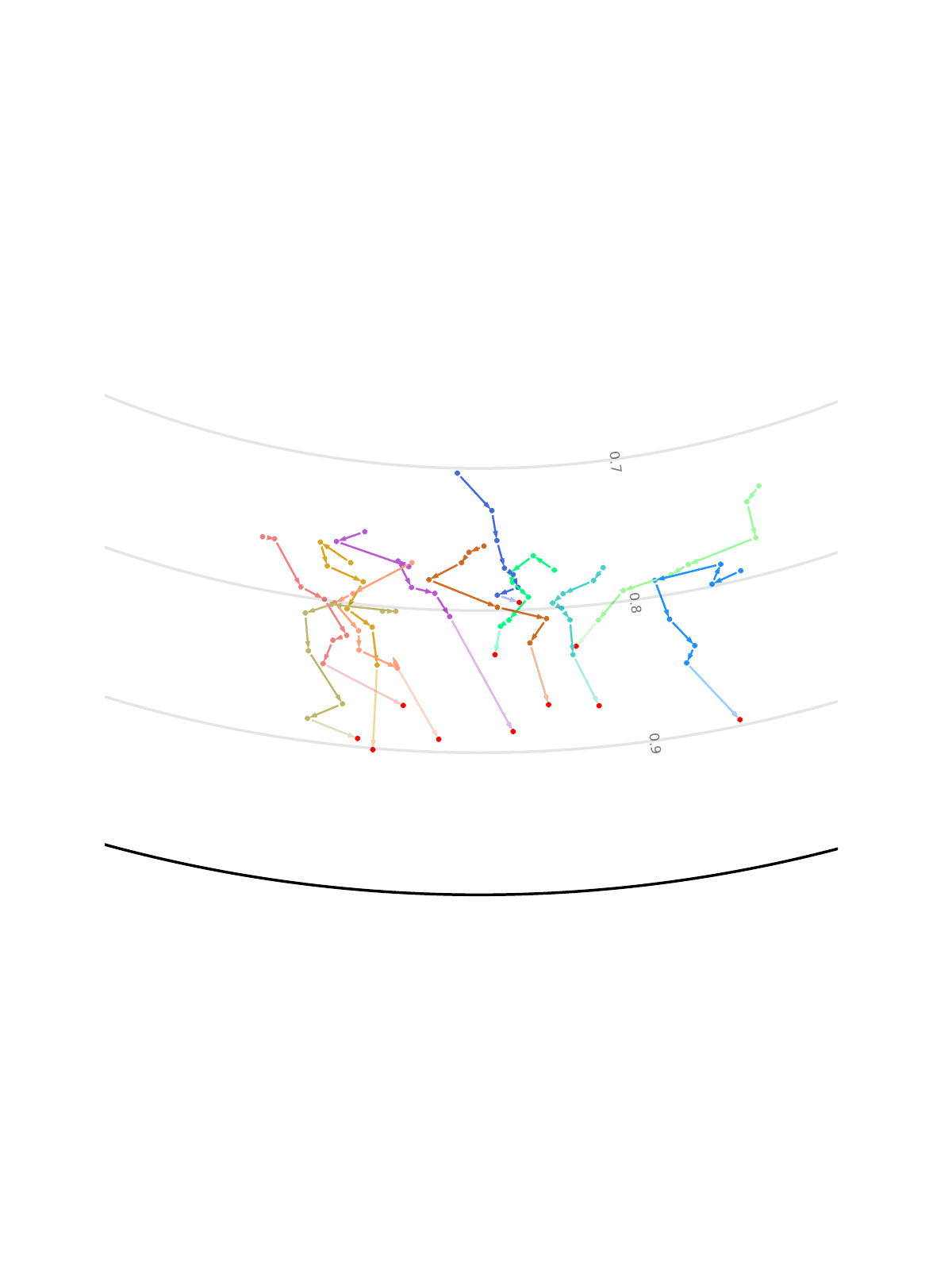}
\caption{Trajectories showing the evolution with time of the predictions in Kinetics, where the task consists in predicting features of the last video clip. We show the two dimensions with larger values across all the predictions. Each line represents a prediction for a specific video.}
\label{fig:trajectories}
\vspace{-3mm}
\end{figure}


\subsection{Evaluation metrics}
In our quantitative evaluation, our goal is twofold. First, we want to evaluate that the obtained representations are better at modeling scenarios with high uncertainty. Second, we want to analyze the hierarchical structure of the learned space. We use three different metrics:

\textbf{Accuracy:} Standard classification accuracy. Compares only classes at the lowest (most specific) hierarchical level. 

\textbf{Bottom-up hierarchical accuracy:} A prediction is considered partially correct if it predicts the wrong node at the leaf level but the correct nodes at upper levels. We weigh each level with a reward that decays by 50\% as we go up in the hierarchy tree.

\textbf{Top-down hierarchical accuracy:} In the hierarchical metrics literature it is sometimes argued that predicting the root node correctly is more important than predicting the exact leaf node \cite{wu2019hierarchical}. Therefore we also report the accuracy value that gives the root node a weight of 1, and decreases it the closer we get to the leaf node, also by a factor of 1/2.

For clarity, in both hierarchical evaluations we normalize the accuracy to be always within the $[0, 1]$ range. In all cases, higher is better.


\subsection{Baselines}
The main goal of the paper is to compare \textbf{hyperbolic} representations to \textbf{Euclidean} ones \cite{han2019video}. We therefore present our experiments on comparisons between these two spaces, keeping the rest of the method the same. It is worth noting that \cite{han2019video} has state-of-the-art results for several video tasks among self-supervised video approaches. Additionally, we report \textbf{chance} accuracy, resulting from randomly selecting a class, and the \textbf{most common} strategy, which always selects the most common class in the training set. 

Trees are compact in hyperbolic space.
We show results for feature spaces with $256$ dimensions, as in \cite{han2019video}, as well as $64$ dimensions.

\subsection{Early Action Recognition}
\label{sec:early_action}

We first evaluate on the early action recognition task, which aims to classify actions that have started, but not finished yet. In other words, the observed video is only partially completed, producing uncertainty. We use video-level action labels to train the classification layer on $\hat{z}_{N}(c_t)$, for all time steps $t$. Tab.~\ref{tab:earlyaction_finegym} and Tab.~\ref{tab:earlyaction_hollywood2} show that, with everything else fixed, hyperbolic models learn significantly better representations than the Euclidean counterparts (up to 14\% gain). The hyperbolic representation enjoys substantial compression efficiency, indicated by the $64$ dimensional hyperbolic embedding outperforming the larger $256$ dimensional Euclidean embedding (up to 5\%). As indicated by both hierarchical accuracy metrics, when there is uncertainty, the hyperbolic representation will predict a more appropriate parent than Euclidean representations.



\begin{table}[t]
\centering
\small
\begin{tabular}{l c | c}
\toprule
\textbf{Representation} & \textbf{Dim.} & \textbf{Accuracy (\%)} \\
\hline
Hyperbolic (ours) & 256  & \textbf{82.54} \\ 
Euclidean \cite{han2019video} & 256 & 68.16 \\ 
Hyperbolic  (ours)  & 64  & 73.35 \\ 
Euclidean \cite{han2019video} & 64   & 66.04 \\ 
Most common &  & 35.40 \\
Chance &  & 25.00 \\
\bottomrule
\end{tabular}
\caption{Early action prediction on FineGym. We do not report hierarchical accuracy because FineGym only annotates the hierarchy at the clip level, not video level. See Section~\ref{sec:early_action} for discussion.}
\vspace{1em}
\label{tab:earlyaction_finegym}
\centering
\small
\begin{tabular}{l c | c c c}
\toprule
\textbf{Representation} & \textbf{Dim.} & \textbf{Accuracy} & \makecell{\textbf{Top-down} \\ \textbf{hier. acc.}} & \makecell{\textbf{Bottom-up} \\ \textbf{hier. acc.}} \\
\hline
Hyperbolic (ours)  & 256 & \textbf{23.10} & \textbf{33.99} & \textbf{28.55} \\ 
Euclidean \cite{han2019video} & 256 & 21.77 & 33.00 & 27.38 \\ 
Hyperbolic (ours) & 64  & 22.25 & 31.47 & 26.86 \\ 
Euclidean \cite{han2019video} & 64  & 15.47 & 24.09 & 19.78 \\ 
Most common & & 17.08 & 25.19 & 21.13  \\
Chance & & 8.33 & 16.11 & 12.22  \\
\bottomrule
\end{tabular}
\caption{Early action prediction on Hollywood2. See Section~\ref{sec:early_action}.}
\label{tab:earlyaction_hollywood2}
\vspace{-1em}
\end{table}


\begin{figure}
    \centering
    \includegraphics[width=\linewidth]{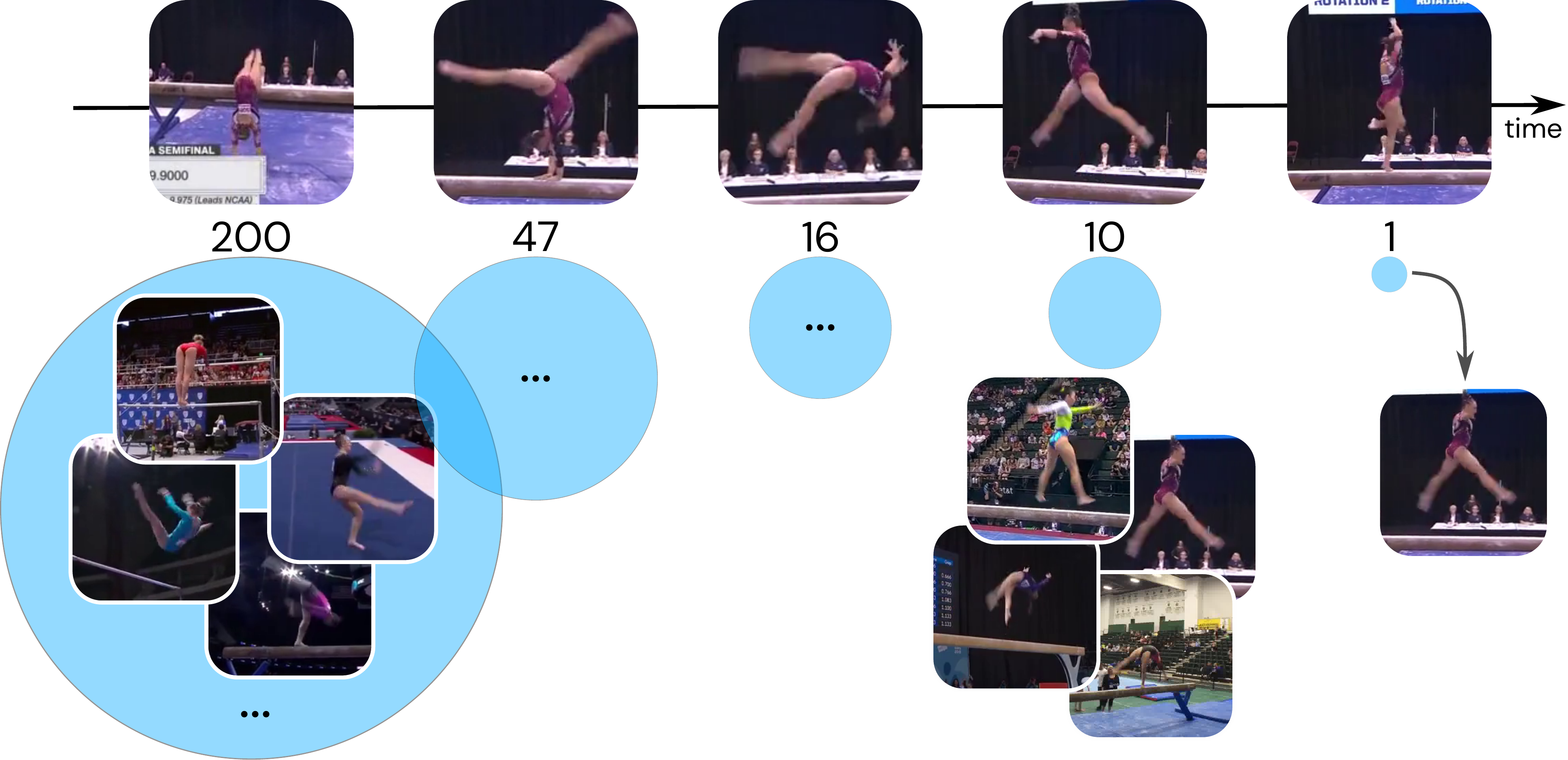}
    \caption{The area of the circle is proportional to the number of videos retrieved within a certain threshold distance. The more specific the prediction gets, the further most of the clips are (only a few ones get closer). The threshold is computed as the mean of all the distances from predictions to clip features. Note that the last circle does not necessarily have to contain exactly one video clip. The total number of clips for this retrieval experiment was 300.}
    \label{fig:retrieved_time}
    \vspace{1em}
    \includegraphics[width=\linewidth]{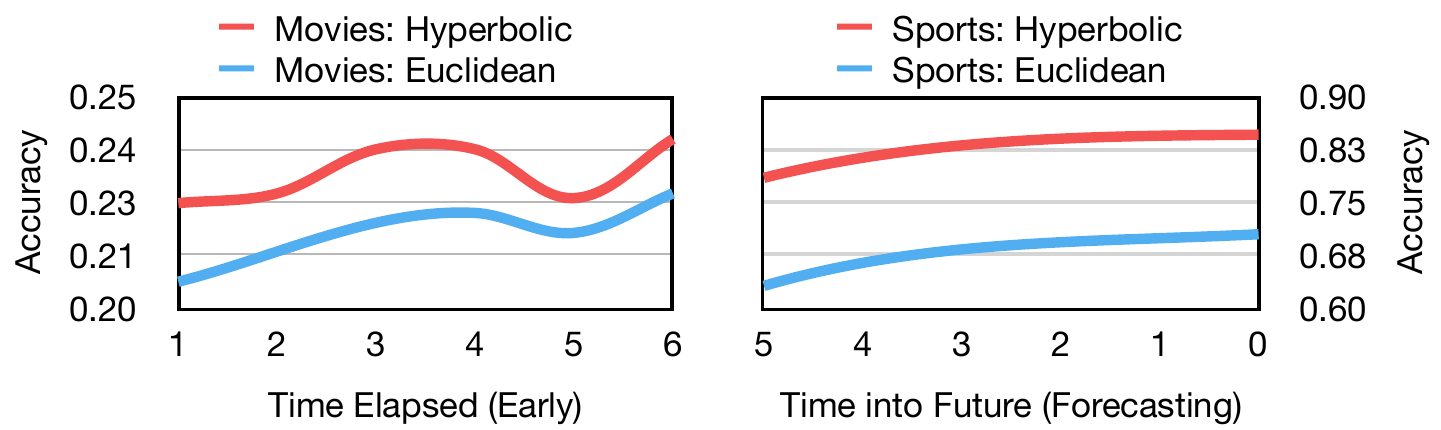}
    \caption{Classification accuracy for early action prediction (left) and future action prediction (right). Performance increases as the model receives more observations or predicts closer to the target. For all horizons, the hyperbolic representation is more accurate.}
    \label{fig:eta}
    \vspace{-1em}
\end{figure}

\subsection{Future Action Prediction}
\label{sec:future_action}

We next evaluate the representation on future action prediction, which aims to predict actions before they start given the past context. There is uncertainty because the next actions are not deterministic.

A key advantage of hyperbolic representations is that the model will automatically decide to select the level of abstraction based on its estimate of the uncertainty. If the prediction is closer to the center of the Poincar\'e ball, the model lacks confidence and it predicts a parental node close to the root in the hierarchy. If the prediction is closer to the border of the Poincar\'e ball, the model is more confident and consequently predicts a more specific outcome. 

We fine-tune our model to learn to predict the class of the last clip of a video at each time step, for each of the three hierarchy levels in the FineGym dataset. We use clip-level labels to train the classification layer on the model's prediction $\hat{z}_{N}(c_{t})$. We select a threshold between hierarchy levels by giving each level the same probability of being selected: the predictions that have a radius in the smaller than the 33\% percentile will select the more general level, the ones above the 66\% percentile will select the more specific level, and the rest will select the middle level.\footnote{These thresholds can be modified according to the risk tolerance of the application.} Once the thresholds are set, we obtain both the predicted hierarchy level as well as the predicted class within that level.

Table~\ref{tab:futureaction_finegym}
compares our predictive models versus the baseline in Euclidean space. We report values for $t=N-1$. For all three metrics, predicting a hierarchical representation substantially outperforms baselines by up to $14$ points. The gains in both top-down and bottom-up hierarchical accuracy show that our model selects a better level of abstraction than the baselines in the presence of uncertainty. We visualize the hyperbolic representation and the resulting hierarchical predictions in Fig.~\ref{fig:tree_finegym}. We show more examples of forecasted levels and classes in Fig~\ref{fig:more_examples}.

The hyperbolic model also obtains better performance than the Euclidean model at the standard classification accuracy, which only evaluates the leaf node prediction. Since classification accuracy does not account for the hierarchy, this gain suggests hyperbolic representations help even when the future is certain. We hypothesize this is because the model is explicitly representing uncertainty, which stabilizes the training compared to the Euclidean baseline.

\begin{figure*}[h]
  \vspace{-0.5em}
  \includegraphics[width=\textwidth]{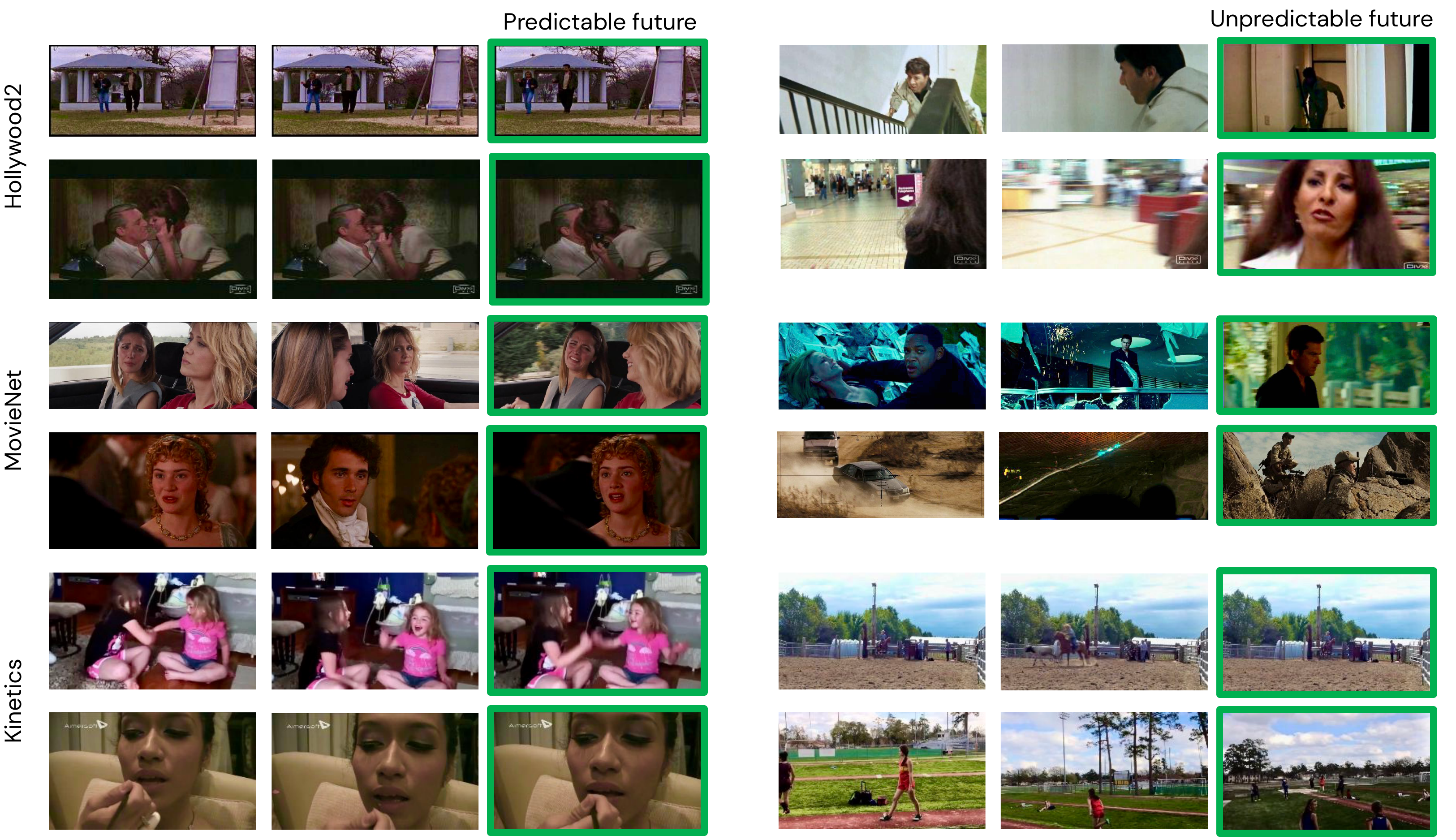}
  \caption{We show examples with high predictability (left), and low predictability (right). The first two frames represent the content the model sees, and the frame in green represents the action the model has to predict. The high predictability examples are selected from above the 99 percentile and the low predictability examples are selected from below the 1 percentile, measured by the radius of the prediction. In both Hollywood2 and MovieNet, the model is more certain about static actions, where the future is close to the past. Specifically in MovieNet we found that most of the highly predictable clips involved close-up conversations of people. The unpredictable ones correspond mostly to action scenes where the future possibilities are very diverse. In the Kinetics dataset, we also noticed that most of the predictable futures corresponded to videos with static people, and the unpredictable ones were associated generally to sports.}
  \label{fig:predictable_examples}
  \vspace{-1em}
\end{figure*}

Since our model represents its prediction of the uncertainty, we are able to visualize which videos are predictable. Fig.~\ref{fig:predictable_examples} visualizes several examples.

\begin{table}[t]
\centering
\small
\begin{tabular}{l c | c c c}
\toprule
\textbf{Representation} & \textbf{Dim.} & \textbf{Accuracy} & \makecell{\textbf{Top-down} \\ \textbf{hier. acc.}} & \makecell{\textbf{Bottom-up} \\ \textbf{hier. acc.}} \\
\hline
Hyperbolic (ours) & 256 & \textbf{13.37} & \textbf{66.64} & \textbf{33.04} \\ 
Euclidean \cite{han2019video} & 256  & 10.08 & 52.00 & 24.75 \\ 
Hyperbolic (ours) & 64  & 10.29 & 56.67 & 27.49 \\ 
Euclidean \cite{han2019video} & 64  & 9.26 & 52.41 & 26.22 \\ 
Most common & & 3.64 & 27.90 & 12.75  \\
Chance & & 0.00 & 16.24 & 5.67  \\
\bottomrule
\end{tabular}
\caption{Future action prediction on FineGym. See Section~\ref{sec:future_action}.}
\label{tab:futureaction_finegym}
\vspace{-1em}
\end{table}

\begin{figure}
    \centering
    \includegraphics[width=\linewidth]{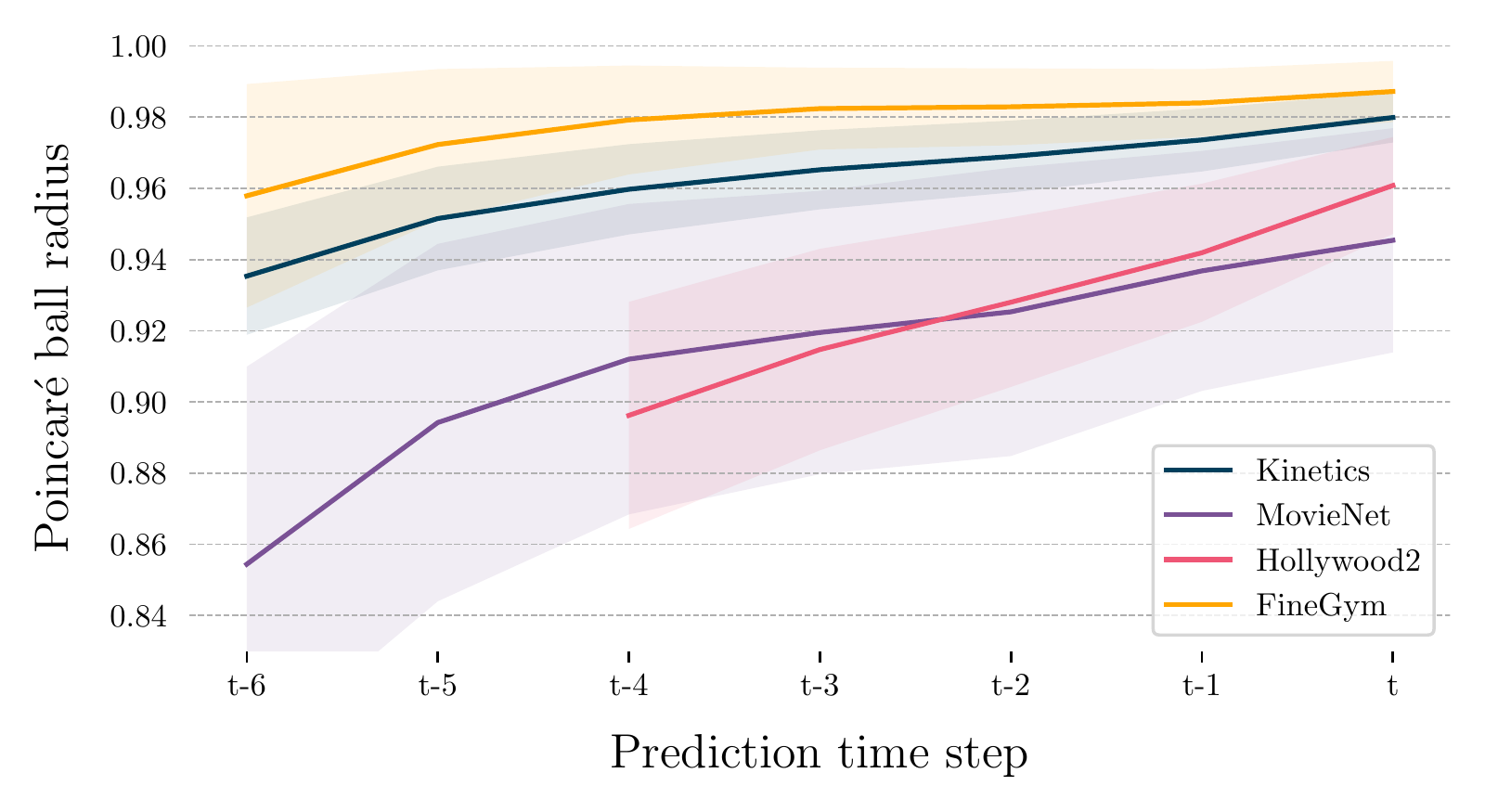}
    \caption{We visualize how the radius of predicted features evolve as more data is observed. The larger the radius, the more confident and more specific the prediction becomes.}
    \label{fig:earlyaction_radius}
    \vspace{-1em}
\end{figure}

\subsection{Analysis of the Representation}

We next analyze the emergent properties of the learned representations, and how they change as more information is given to the model. We conduct our analysis on the self-supervised representation, before supervised fine-tuning. 


Fig.~\ref{fig:trajectories} visualizes the trajectory that representations follow as frames are continuously observed. We visualize the representation from Kinetics. In order to plot a 2D graph, we select the two dimensions with the highest mean value, and plot their trajectories with the time.\footnote{Projecting using TSNE \cite{maaten2008visualizing} or uMAP \cite{mcinnes2018umap} does not respect the local structure well enough to visualize the radius of the prediction.} The red dots show the actual features that have to be predicted. As the observations get closer to its prediction target, predictions get closer to the edge of the Poincar\'e ball, indicating they are becoming more specific in the hierarchy and increasingly confident about the prediction.


The distance to the center of the Poincar\'e ball gives us intuition about the underlying geometry, and
Fig.~\ref{fig:earlyaction_radius} quantifies this behavior. We show the average radius of the predictions at each time step, together with the standard deviation. As more frames become available, the prediction gets consistently more confident for all datasets.

Abstract predictions will encompass a large number of specific features that can be predicted, while specific predictions will restrict the options to just a few. We visualize these predictions using nearest neighbors.
Given a series of video clips belonging to the same event and gymnastics instrument in FineGym, we compute features for each one of the clips in these videos, that we use as targets to retrieve. For each video, we then predict the last representation at each time step (i.e., for every clip in the video), and use these as queries. We show the results for one such videos in Fig.~\ref{fig:retrieved_time}. The retrieved number corresponds to the number of clips that are in a distance within a threshold, that we compute as the mean of all the distances from predictions to features. As the time horizon to the target action shrinks, the more specific the representation becomes, and thus fewer options are recalled. Fig.~\ref{fig:eta} quantifies performance versus time horizon, showing the hyperbolic representation is more accurate than the Euclidean representation for all time periods. 





\vspace{-0.5em}
\section{Conclusion}
\vspace{-0.5em}

While there is uncertainty in the future, parts of it are predictable. We have introduced a hyperbolic model for video prediction that represents uncertainty hierarchically. After learning from unlabeled video, experiments and visualizations show that a hierarchy automatically emerges in the representation, encoding the predictability of the future.  

{
\small
\textbf{Acknowledgments:} We thank Will Price, Mia Chiquier, Dave Epstein, Sarah Gu, and Ishaan Chandratreya for helpful feedback. This research is based on work partially supported by NSF NRI Award \#1925157, the DARPA MCS program under Federal Agreement No.\ N660011924032, the DARPA KAIROS program under PTE Federal Award No.\ FA8750-19-2-1004, and an Amazon Research Gift. We thank NVidia for GPU donations. The views and conclusions contained herein are those of the authors and should not be interpreted as necessarily representing the official policies, either expressed or implied, of the U.S. Government.
}

{\small
\bibliographystyle{ieee_fullname}
\bibliography{egbib}
}

\appendix
\clearpage
\section*{Appendix}

\section{Hollywood2 Hierarchy}
\label{sec:appendix_hollywood}

\begin{figure}[h]
\includegraphics[width=0.7\linewidth]{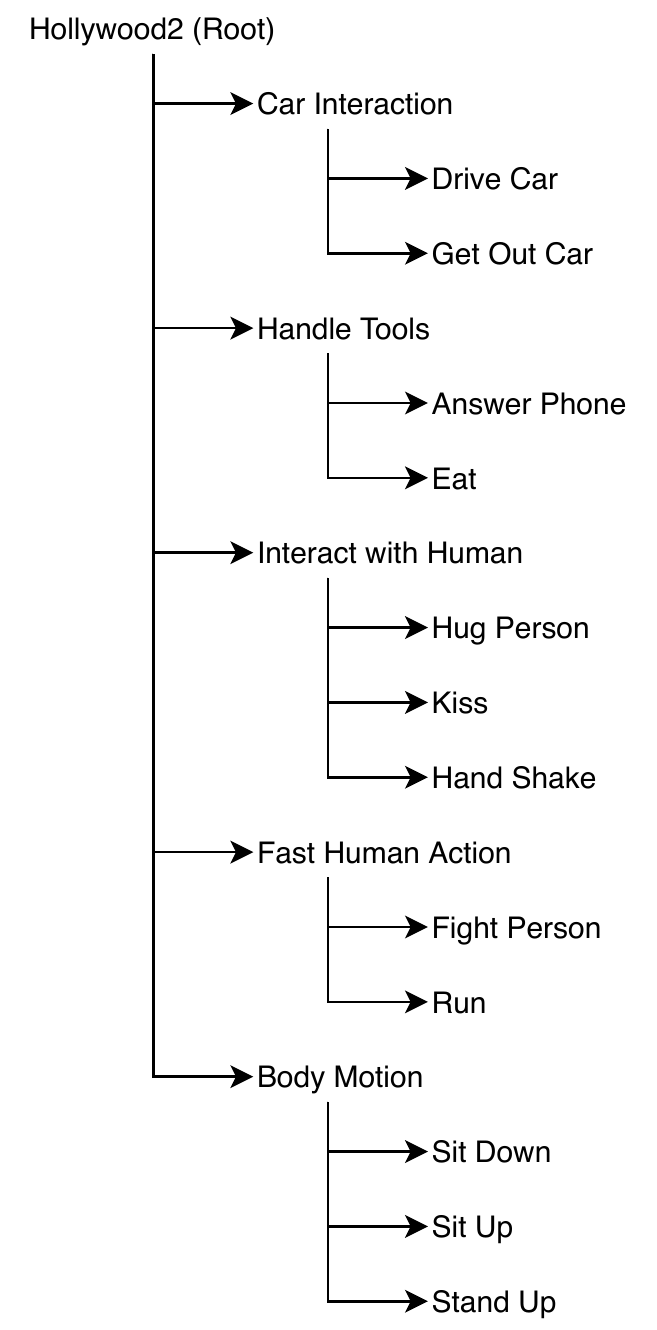}
\end{figure}

\section{Implementation Details}
\label{sec:appendix_impl_details}

Code and models are publicly available on \href{https://github.com/cvlab-columbia/hyperfuture}{github.com/cvlab-columbia/hyperfuture}. 

We implement $f$ with a 3D-ResNet18, and $g$ as a one-layer Convolutional Gated Recurrent Unit (ConvGRU) with kernel size $(1,1)$. We implement $\phi$ using a two-layer perceptron. The dimensionality at the output of the ResNet is 256. When training with representations with smaller dimensionality, we add an extra linear layer to project the representations.

Before inputting representations to the classification layer, we mean pool them spatially, in the Euclidean space.

In order to select video clips, we subsample $N$ non-overlapping blocks from the original video, consisting of $T$ frames each. In cases where divisions into clips are given by the dataset, we use those as inputs. By default, we use $T=5$ except for datasets where clips are shorter than 5 frames, in which case we use the size of the given clip. Similarly, we set $N=8$ except in datasets where videos are short, in which case we use the maximum size allowed by the dataset. Following \cite{han2019video} we use $H=W=128$.

We optimize our objective using Riemannian Adam \cite{becigneul2018riemannian} and the Geoopt library in Pytorch \cite{geoopt2020kochurov,pytorch}. We train our models with a batch size of 128 for 100 epochs. We tried three learning rates ($10^{-2}$, $10^{-3}$, $10^{-4}$). We select the model with better accuracy on a held-out validation set. 

We select positives and negatives using the same strategy as \cite{han2019video}, augmenting it with $\delta$ values larger than 1.

\typeout{get arXiv to do 4 passes: Label(s) may have changed. Rerun}
\end{document}